\documentclass[letterpaper]{article} 
\usepackage{aaai25}  
\usepackage{times}  
\usepackage{helvet}  
\usepackage{courier}  
\usepackage[hyphens]{url}  
\usepackage{graphicx} 
\usepackage{natbib}  
\usepackage{caption} 
\usepackage{algorithm}
\usepackage{algorithmic}
\usepackage{amsmath}
\usepackage{amsthm}
\usepackage{amssymb}
\usepackage{tcolorbox}
\usepackage{booktabs}
\usepackage{multirow}
\newtheorem{definition}{Definition}

\usepackage{newfloat}
\usepackage{listings}

\definecolor{nicered}{rgb}{.617, .039, .031}
\definecolor{nicegreen}{rgb}{.023, .242, .168}
\definecolor{nicegreen}{rgb}{0, .390625, 0}
\definecolor{niceblue}{rgb}{0.125, 0.406, 0.852}

\DeclareCaptionStyle{ruled}{labelfont=normalfont,labelsep=colon,strut=off} 
\floatstyle{ruled}
\newfloat{listing}{tb}{lst}{}
\floatname{listing}{Listing}
\definecolor{niceblue}{rgb}{0.125, 0.406, 0.852}

\title{Achieving the Tightest Relaxation of Sigmoids for Formal Verification}

\author {
    Samuel Chevalier\textsuperscript{\rm 1},
    Duncan Starkenburg\textsuperscript{\rm 1},
    Krishnamurthy (Dj) Dvijotham\textsuperscript{\rm 2}
}
\affiliations {
   \textsuperscript{\rm 1}University of Vermont\\
   \textsuperscript{\rm 2}ServiceNow\\
   samuel.chevalier@uvm.edu, duncan.starkenburg@uvm.edu, dvij@cs.washington.edu}

\usepackage{bibentry}

\begin{document}

\maketitle

\begin{abstract}

In the field of formal verification, Neural Networks (NNs) are typically reformulated into equivalent mathematical programs which are optimized over. To overcome the inherent non-convexity of these reformulations, convex relaxations of nonlinear activation functions are typically utilized. Common relaxations (i.e., static linear cuts) of ``S-shaped" activation functions, however, can be overly loose, slowing down the overall verification process. In this paper, we derive tuneable hyperplanes which upper and lower bound the sigmoid activation function. When tuned in the dual space, these affine bounds smoothly rotate around the nonlinear manifold of the sigmoid activation function. This approach, termed $\alpha$-sig, allows us to tractably incorporate the tightest possible, element-wise convex relaxation of the sigmoid activation function into a formal verification framework. We embed these relaxations inside of large verification tasks and compare their performance to LiRPA and $\alpha$-CROWN, a state-of-the-art verification duo.

\end{abstract}

%

\section{Introduction}
Formal verification has an ever-widening spectrum of important uses, including mathematical proof validation~\cite{trinh2024solving}, adversarially robust classification~\cite{adv_attack}, data-driven controller reachability analysis~\cite{everett2021neural}, performance guarantees for surrogate models of the electric power grid~\cite{chevalier2024global}, and more~\cite{urban2021reviewformalmethodsapplied}. The formal verification of Neural Networks (NNs), in particular, has seen a flurry of recent research activity. Pushed by the international Verification of Neural Networks Competition (VNN-Comp), NN verification technologies have scaled rapidly in recent years~\cite{brix2023years,brix2023fourth}. Competitors have exploited, and synergistically spurred, the development of highly successful verification algorithms, e.g., $\alpha,\beta$-CROWN~\cite{wang2021beta,lyu2019fastened}, Multi-Neuron Guided Branch-and-Bound~\cite{ferrari2022complete}, DeepPoly~\cite{singh2019abstract}, etc. The winningest methods emerging from VNN-Comp serve as the leading bellwethers for state-of-the-art within the NN verification community. 

Despite these advances, verification technologies cannot yet scale to Large Language Model (LLM) sized systems~\cite{sun2024trustllmtrustworthinesslargelanguage}. Nonlinear, non-ReLU activation functions present one of the key computational obstacles which prevents scaling. While these activation functions can be attacked with spatial Branch-\&-Bound (B\&B) approaches~\cite{shi2024neural}, authors in ~\cite{lazy_s_shape} note that ``existing verifiers cannot tightly approximate S-shaped activations." The sigmoid activation is one such S-shaped activation function which is challenging to deal with. Given its close relationship to the ubiquitous softmax function~\cite{wei2023convexboundssoftmaxfunction}, which is embedded in modern transformer layers~\cite{ildiz2024selfattentionmarkovmodelsunveiling}, efficient verification over the sigmoid activation function would help boost verification speeds and generally help extend verification technology applicability.

\textbf{Our contributions.} Given the ongoing computational challenge of verifying over NNs containing sigmoid activation fucntions, our contributions follow:
\begin{enumerate}
    \item We derive an explicit mapping between the linear slope and y-intercept point of a tangent line which tightly bounds a sigmoid. This differentiable, tunable mapping is embedded into a verification framework to yield the tightest possible element-wise relaxation of the sigmoid.
    \item We propose a ``backward" NN evaluation routine which dynamically detects if a sigmoid should be upper or lower bounded at each step of a gradient-based solve.
    \item To ensure feasible projection in the dual space, we design a sequential quadratic program which efficiently pre-computes maximum slope bounds of all tunable slopes.
    
\end{enumerate}

Many verification algorithms exploit element-wise convex relaxation of nonlinear activation functions, resulting in convex solution spaces~\cite{salman2020convexrelaxationbarriertight}. The stacking up of tightening dual variables within a dualized problem reformulation, as in $\alpha,\beta$-CROWN, can lead to a formally \textit{nonconvex} mathematical program. Similarly, the verification formulation which we present in this paper is nonconvex, but it globally lower bounds the true verification solution. Authors in~\cite{bunel2020efficientnonconvexreformulationstagewise} have shown that spurious local minima in nonconvex problems containing ``staged convexity" are very rare, and they even design perturbation-based approaches to avoid them. In this paper, we use a gradient-based approach to solve a formally nonconvex problem, but solutions smoothly converge to what appears to be a global maximum. 

\section{Related Works}
To iteratively tighten relaxed ReLU-based NNs, the architects of $\alpha$-CROWN~\cite{xu2021fast} use the ``optimizable" linear relaxation demonstrated in Fig.~\ref{ref:relu}. In this approach, the parameter $\alpha$ is tuned to achieve a maximally tight lower bound on the verification problem. Authors in~\cite{salman2020convexrelaxationbarriertight} showed $\alpha$ variables to be equivalent to the dual variables of an associated linear program (LP)-relaxed verification problem.~\cite{wang2021beta} applied the $\alpha$-CROWN approach within a B\&B context, introducing new split-constraint dual variables $\beta$ to be optimized over in the dual space. More recently,~\cite{shi2024neural} developed a general framework (GenBaB) to perform B\&B over a wide class of nonlinear activations, including sigmoid, tanh, sine, GeLU, and bilinear functions. The authors utilize pre-optimized branching points in order to choose linear cuts which statically bound portions of the activation functions. Critically, they also incorporate optimizable linear relaxations of the bilinear, sine, and GeLU activation functions using $\alpha$-like tunable parameters.

Verification over ``S-shaped" activation functions is considered in~\cite{lazy_s_shape}, where the authors use sequential identification of counterexamples in the relaxed search space to iteratively tighten the activation relaxations. Authors in~\cite{zhang2022provablytightestlinearapproximation} propose a verification routine which incorporates the provably tightest linear approximation of sigmoid-based NNs. Notably, the approach uses static (i.e., non-tunable) linear approximations, improving upon other works whose sigmoid relaxations minimize relaxation areas~\cite{henriksen2020efficient} or use parallel upper and lower bounding lines~\cite{Wu_Zhang_2021}. While existing approaches have designed advanced cut selection procedures for sigmoid activation functions, and even embedded these procedures within B\&B, there exists no explicitly optimizable, maximally tight linear relaxation strategy for sigmoid activation functions.

\section{Formal Verification Framework}
Let $x\in \mathbb{R}^{n_1}$ be the input to an $L$-layer NN mapping ${\rm NN} (\cdot): \mathbb{R}^{n_1} \rightarrow \mathbb{R}^{n_L}$. A scalar verification metric function, $m(\cdot): \mathbb{R}^{n_L} \rightarrow \mathbb{R}^1$, wraps around the NN to generate the verification function $f(x)\triangleq m({\rm NN}(x))$. This metric is defined such that the NN's performance is \textit{verified} if $f(x)\ge 0$, $\forall x \in \mathcal C$, can be proved:
\begin{align}\label{eq: gamma}
\gamma\triangleq \min_{x\in\mathcal{C}}\; f(x).
\end{align}
In this NN, ${\hat x}^{(i)}= W^{(i)}x^{(i)}+b^{(i)}$ is the $i^{\rm th}$ layer linear transformation, with ${x}^{(1)} = x$ as the input, and ${x}^{(i+1)}=\sigma({\hat x}^{(i)})$ is the associated nonlinear activation. In this paper, $\sigma(\cdot)$ exclusively represents the sigmoid activation function:
\begin{align}
\sigma(x)\triangleq\frac{e^{x}}{1+e^{x}}=\frac{1}{1+e^{-x}},
\end{align}
where the gradient of the sigmoid is $\sigma'=\sigma(1-\sigma)$. As in~\cite{wang2021beta}, the region $\mathcal C$ can be described as an $\ell_p$ norm ball constraint on the input via $\mathcal{C}=\{x\,|\,|| x-x_0||_{p}\le\epsilon\}$. To solve \eqref{eq: gamma} to global optimality (or, at least, to prove $\gamma \ge 0$), many recent approaches have utilized ($i$) convex relaxation of activation functions coupled with ($ii$) spatial~\cite{shi2024neural} and discrete~\cite{wang2021beta} Branch-and-Bound (B\&B) strategies. B\&B iteratively toggles activation function statuses, yielding tighter and tighter solutions, in pursuit of the problem's true lower bound. In this paper, we exclusively consider the so-called root node relaxation of \eqref{eq: gamma}, i.e., the first, and generally loosest, B\&B relaxation, where all activation functions are simultaneously relaxed. We denote the root node relaxation of $f(x)$ as ${\tilde f}_r(x)$, where
\begin{align}\label{eq: min_fr}
\min_{x\in\mathcal{C}}\;f(x)\;\ge\;\min_{x\in\mathcal{C}}\;\tilde{f}_{r}(x).
\end{align}
is guaranteed, assuming valid relaxations are applied. While $\tilde{f}_{r}(x)$ is an easier problem to solve,~\cite{salman2020convexrelaxationbarriertight} showed that verification over relaxed NNs can face a ``convex relaxation barrier" problem; essentially, even tight relaxations are sometimes not strong enough to yield conclusive verification results. In this paper, we seek to find the tightest possible relaxation of the sigmoid activation function. 

\section{Sigmoid Activation Function Relaxation}

\begin{figure}[t]
\centering
\includegraphics[width=0.99\columnwidth]{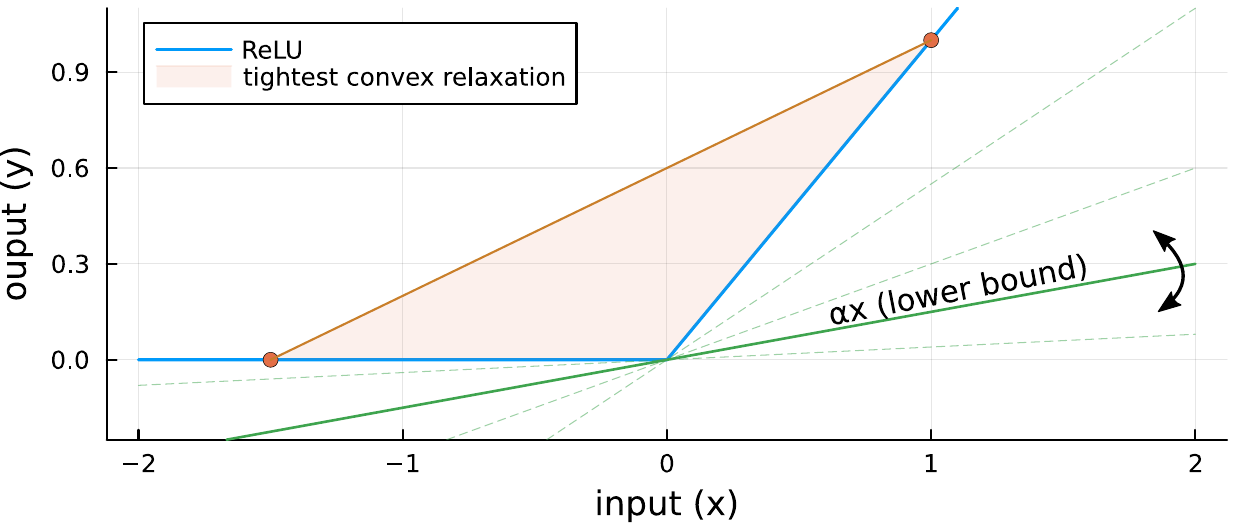}
\caption{Convex relaxation of ReLU activation function, lower bounded by $y=\alpha x$, with tunable $\alpha\in [0,1]$. }
\label{ref:relu}
\end{figure}

In order to convexly bound the ${\rm ReLU}(x)={\rm max}(x,0)$ activation function, \cite{xu2021fast} famously replaced the tight ``triangle" LP-relaxation of an unstable neuron with a tunable lower bound, $\alpha$, as illustrated in Fig.~\ref{ref:relu}. This $\alpha$ was then iteratively maximized over in the dual space to achieve the tightest lower bound of the relaxed NN. At each gradient step, the value of $\alpha$ was feasibly clipped to $0\le \alpha \le 1$, such that  $\alpha x \le {\rm ReLU}(x)$ was always maintained.

In the same spirit, we seek to bound the sigmoid activation function $\sigma (x)$ with tunable affine expressions via 
\begin{align}
\alpha_l x + \beta_l \;\le\; \sigma(x) \;\le\; \alpha_u x + \beta_u,
\end{align}
where $\alpha_l$, $\beta_l$, $\alpha_u$, and $\beta_u$ are maximized over in the dual space to find the tightest lower bound on the NN relaxation. Analogous to the $0\le \alpha \le 1$ bound from \cite{xu2021fast}, however, we must ensure that the numerical values of $\alpha_l$, $\beta_l$, and $\alpha_u$, $\beta_u$ always correspond to, respectively, valid lower and upper bounds on the sigmoid activation function. 

\begin{definition}
    In this paper, 
    \begin{itemize}
    \item $\alpha_l x + \beta_l$ is called an ``affine \textbf{lower} bound", while 
    \item $\alpha_u x + \beta_u$ is called an ``affine \textbf{upper} bound".
    \end{itemize}
\end{definition}

\begin{figure}[t]
\centering
\includegraphics[width=0.99\columnwidth]{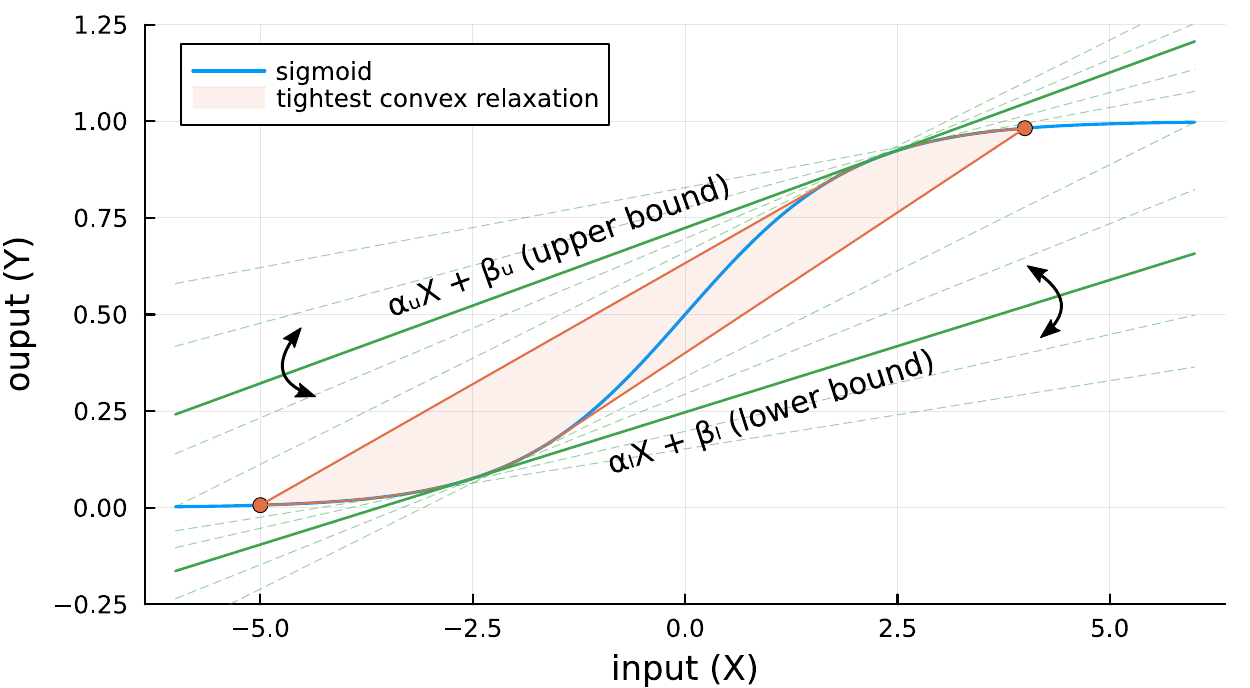}
\caption{Rotating affine bounds around the sigmoid activation function, achieving maximally tight convex relaxation.}
\label{sigmoig_relaxation}
\end{figure}

\begin{figure}[t]
\centering
\includegraphics[width=0.99\columnwidth]{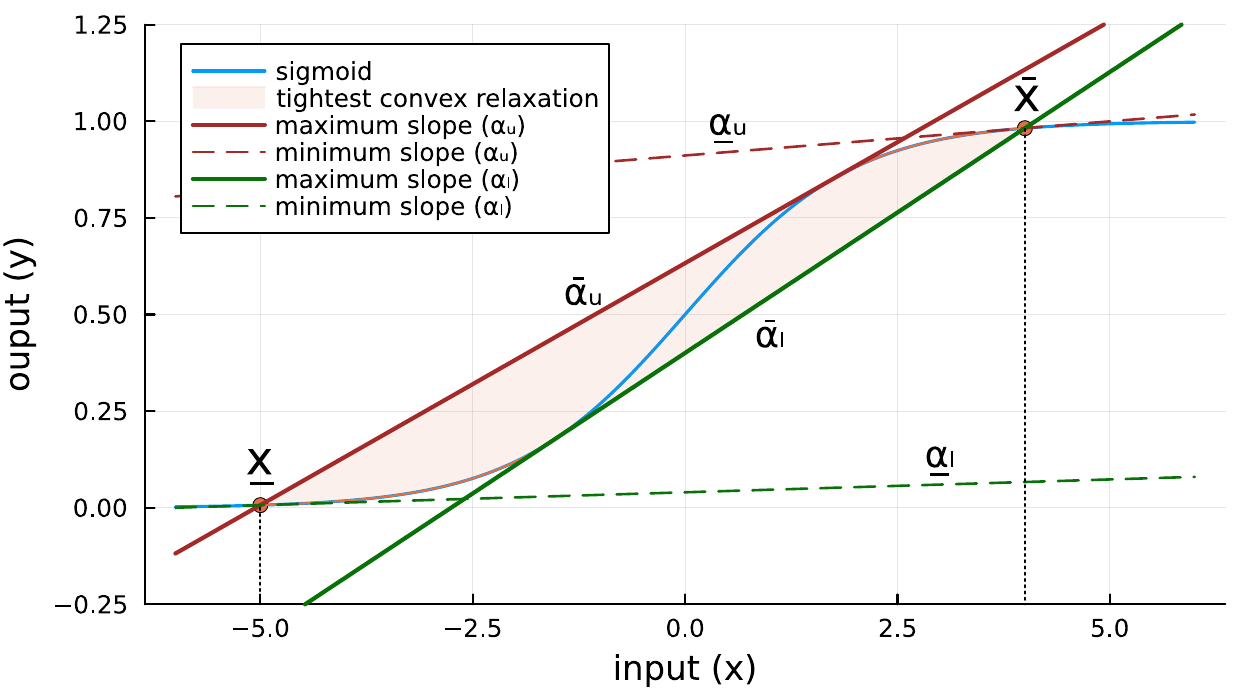}
\caption{For max and min input values ($\overline x$, $\underline x$), depicted are the steepest and shallowest upper bound slopes (${\overline \alpha}_u$, ${\underline \alpha}_u$), and the steepest and shallowest lower bound slopes (${\overline \alpha}_l$, ${\underline \alpha}_l$).}
\label{sig_4_bounds}
\end{figure}

In order to derive the slope $\alpha$ and intercept $\beta$ terms which yield maximally tight convex relaxation of the sigmoid function, we consider the point at which the associated rotating bound $\alpha x + \beta$ intersects with the sigmoid at some tangent point (the upper $u$ and lower $l$ subscripts are dropped for notational convenience). We encode the intersection \eqref{eq: intersect} and tangent \eqref{eq: tangent} point relations via
\begin{subequations}\label{eq: affine_sigmoid_solution}
\begin{align}
\alpha x+\beta & =\sigma(x)\label{eq: intersect}\\
\alpha & =\sigma'(x)\label{eq: tangent},
\end{align}
\end{subequations}
where $\sigma'(x)$ is the gradient of the sigmoid. The system of \eqref{eq: affine_sigmoid_solution} represents two equations and three unknowns. In order to maximize over the $\alpha$ and $\beta$ variables independent of the primal variable $x$, it is advantageous to eliminate $x$ entirely. Interestingly, the solution for $\beta$, written strictly in terms of $\alpha$, has a closed form solution $\beta=h(\alpha)$ (see appendix for derivation). This solution represents a main result from this paper, and it is given by
\begin{tcolorbox}
\noindent\vspace*{-0.35cm}
\begin{align}\label{eq: beta_f_alpha}
\!\!\!\!\!\!\!\!\beta =\!\frac{1}{1+e^{\pm\cosh^{-1}\left(\frac{1}{2\alpha}-1\right)}}\!\pm\alpha\cosh^{-1}\!\left(\frac{1}{2\alpha}\!-1\right)\!.
\end{align}
\vspace*{-0.35cm}
\end{tcolorbox}
\noindent In \eqref{eq: beta_f_alpha}, the $\pm$ terms are negative for the upper bounds, and positive for the lower bounds, as stated in the appendix. Associated affine bounds are plotted in Fig.~\ref{sigmoig_relaxation}. As depicted, these bounds are capable of yielding the tightest possible convex relaxation of the sigmoid activation function. Denoting \eqref{eq: beta_f_alpha} by $\beta=h(\alpha)$, we define a set $\mathcal{S}$ of valid $\alpha,\beta$ values:
\begin{align}
\mathcal{S}=\{\alpha,\beta\;|\;\beta=h(\alpha),\;\underline{\alpha}\le\alpha\le\overline{\alpha}\}.
\end{align}
The minimum and maximum allowable slope values, $\underline{\alpha}$ and $\overline{\alpha}$, are predetermined for every sigmoid activation function. Nominally, $\underline{\alpha}\ge 0$, since $\inf(\sigma' (x))=0$, and $\overline{\alpha}\le 1/4$, since $\sup(\sigma' (x))=0.25$. However, tighter slopes generally exist, based on the sigmoid's minimum and maximum input bounds $\underline x$ and $\overline x$. Fig.~\ref{sig_4_bounds} illustrates a typical situation, where there are distinct slope bounds. In this figure, the minimum slopes (dashed lines) are simply computed as the gradients at the minimum and maximum inputs:
\begin{align}
\underline{\alpha}_{l} & =\sigma'(\underline{x})\\
\underline{\alpha}_{u} & =\sigma'(\overline{x}).
\end{align}
The maximum bounding slopes $\overline{\alpha}_{l}$ and $\overline{\alpha}_{u}$, however, are computed as the lines which intersect the sigmoid at two points: the bounded input anchor points ($\underline x$ and $\overline x$), and a corresponding tangent point. The parallelized calculation of these slopes is a pre-processing step involving sequential quadratic formulate iterations, and the associated procedures are discussed in the appendix.

Fig.~\ref{sig_3_bounds} illustrates an alternative situation, where the maximum and minimum affine upper bound slopes are equal\footnote{Depending on the values of $\underline x$ and $\overline x$, this can also happen to the affine lower \textit{bounds}: $\overline{\alpha}_{l}=\underline{\alpha}_{l}$. However, due to the nonlinearity of the sigmoid, $\overline{\alpha}_{l}=\underline{\alpha}_{l}$ and $\overline{\alpha}_{u}=\underline{\alpha}_{u}$ cannot occur simultaneously, unless ${\overline x}={\underline x}$, which is a degenerate case.}: $\overline{\alpha}_{u}=\underline{\alpha}_{u}$. This occurs when the tangent slope at $\overline x$ can be ``raised up" such that the corresponding affine bound eventually intersects with $\underline x$. Defining $\Delta x= \overline{x}-\underline{x}$, we have two possibilities:
\begin{subequations}
\begin{align}
 \text{if}\;\;\overline{x}-\sigma'(\overline{x})\Delta x\le\underline{x}, & \;\; \text{then}\;\;\overline{\alpha}_{u}=\underline{\alpha}_{u}\label{eq: upper_equal}\\
 \text{if}\;\;\overline{x}-\sigma'(\overline{x})\Delta x>\underline{x}, & \;\; \text{then}\;\;\overline{\alpha}_{u}>\underline{\alpha}_{u}.
\end{align}
\end{subequations}
Similarly, in the lower affine bound case, 
\begin{subequations}
\begin{align}
 \text{if}\;\;\underline{x}+\sigma'(\underline{x})\Delta x\ge\overline{x}, & \;\; \text{then}\;\;\overline{\alpha}_{l}=\underline{\alpha}_{l}\\
 \text{if}\;\;\underline{x}+\sigma'(\underline{x})\Delta x<\overline{x}, & \;\; \text{then}\;\;\overline{\alpha}_{l}>\underline{\alpha}_{l}.
\end{align}
\end{subequations}
In either case, equal slopes can be directly computed as
\begin{align}
\underline{\alpha}_{u}=\overline{\alpha}_{l}=\frac{\sigma(\overline{x})-\sigma(\underline{x})}{\overline{x}-\underline{x}}.
\end{align}



\begin{figure}[t]
\centering
\includegraphics[width=0.99\columnwidth]{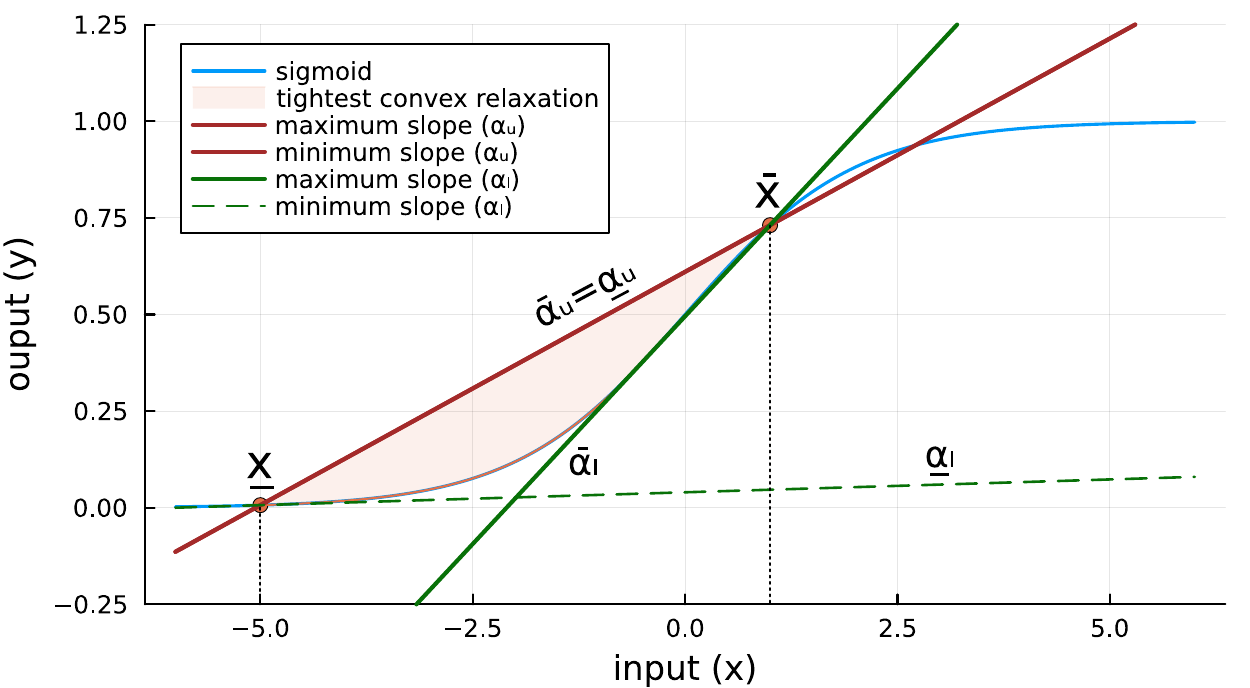}
\caption{Since the slope at $\overline x$ is sufficiently steep, \eqref{eq: upper_equal} is applied, and the tightest upper bound is a static line.}
\label{sig_3_bounds}
\end{figure}

\section{Backward Bound Propagation}
To efficiently minimize $\tilde{f}_{r}(x)$, as in \eqref{eq: min_fr}, we utilize the backward bound propagation procedure employed in, e.g.,~\cite{wang2021beta}. In applying this procedure, however, we utilize a sequential backward evaluation step in order to dynamically detect if a sigmoid function should be upper, or lower, bounded by the affine bound in \eqref{eq: beta_f_alpha}. Consider, for example, the simple problem
\begin{align}
\min_{x\in\mathcal{C}}\;c^{T}\sigma(x)\;\ge\;\min_{x\in\mathcal{C}}\;c^{T}(\alpha x+\beta),
\end{align}
where $c$ is some cost vector. To achieve minimum cost,
\begin{itemize}
    \item if $c_i\ge 0$, then $\alpha_l x +\beta_l$ should lower bound the sigmoid;
    \item if $c_i\le 0$, then $\alpha_u x +\beta_u$ should upper bound the sigmoid.
\end{itemize}
This procedure is sequentially applied as we move backward through the NN layers. The coefficients in front of each layer, however, will be a function of the numerical $\alpha$ values, which will be changing at each gradient step during the verification solve. In the appendix, we define a NN mapping \eqref{eq: nn_mapping}, and then we sequentially move backward through this mapping in \eqref{eq: sign_funcs}, inferring the sign of the coefficients in front of each affine bound term.

\subsection{Dual Verification} Using the affine relaxed version of the NN mapping in \eqref{eq: nn_mapping}, we may start with input $x$ to sequentially replace all intermediate primal variables:
\begin{subequations}
\begin{align}
\tilde{f}_{r}(x) & =c^{T}(...{\rm d}(\alpha^{(i)}\!)(W^{(i)}(...\,x\,...)\!+\!b^{(i)})\!+\!\beta^{(i)}...)\\
 & \triangleq g_{1}(\alpha,\beta)^{T}x+g_{2}(\alpha,\beta).
\end{align}
\end{subequations}
The associated minimization problem over this relaxed verification problem is given by
\begin{subequations}
\begin{align}
\min_{x\in\mathcal{C}}\;\tilde{f}_{r}(x) & =\min_{\left\Vert x\right\Vert _{p}\le\epsilon}\;g_{1}(\alpha,\beta)^{T}x+g_{2}(\alpha,\beta)\\
 & =\;\underbrace{-\left\Vert g_{1}(\alpha,\beta)^{T}\right\Vert _{q}+g_{2}(\alpha,\beta)}_{{\tilde f}_r(\alpha,\beta)},
\end{align}
\end{subequations}
where the dual norm~\cite{wang2021beta,chevalier2024gpu} has been used to transform the $p$ norm constraint into a $q$ norm objective term. Any valid (i.e., feasible) set of $\alpha$, $\beta$ parameters will yield a valid lower bound for the relaxed verification problem. To achieve the \textit{tightest} lower bound, we may maximize ${\tilde f}_r(\alpha,\beta)$ over the feasible set $\mathcal S$ of $\alpha$, $\beta$:
\begin{align}\label{eq: dual}
{\tilde \gamma} = \max_{\{\alpha,\beta\}\in\mathcal{S}}\;-\left\Vert g_{1}(\alpha,\beta)^{T}\right\Vert _{q}+g_{2}(\alpha,\beta).
\end{align}
While $\alpha,\beta$ are not dual variables in the traditional sense, they are responsible for actively constraining the primal space, so we we refer to \eqref{eq: dual} as a dual problem.

While \eqref{eq: dual} can be solved via projected gradient ascent, we may alternatively use $\beta=h(\alpha)$ in order to eliminate the $\beta$ variable entirely. Initial testing shows that eliminating $\beta$, and then backpropagating through $h(\alpha)$, is more effective than equality projecting feasible, via \eqref{eq: beta_f_alpha}, at each step. The updated formulation is given via 
\begin{align}\label{eq: dual_update}
{\tilde{\gamma}}=\max_{{\underline \alpha}\le \alpha\le {\overline \alpha}}\;-\left\Vert g_{1}(\alpha,h(\alpha))^{T}\right\Vert _{q}+g_{2}(\alpha,h(\alpha)).
\end{align}

\subsection{Projected gradient-based solution routine} 
We use a projected gradient routine in order to solve \eqref{eq: dual_update} and enforce ${\underline \alpha}\le \alpha\le {\overline \alpha}$. Sigmoid functions may need to be upper bounded via \eqref{eq: beta_u}, and then lower bounded via \eqref{eq: beta_l}, as numerical values of $\alpha$ evolve. To overcome this challenge, at each step of our numerical routine, we reverse propagate compute the sign vector $s^{(i)}$, from \eqref{eq: sign_funcs}, for each layer. Since this vector tells us if the sigmoid function should be upper or lower bounded, we embed corresponding elements of this vector inside of \eqref{eq: beta_f_alpha}, replacing the $\pm$ terms. At the $j^{\rm th}$ activation function of each $i^{\rm th}$ NN layer, the corresponding slope and sign elements are feasibly related via
{\small\begin{align}\label{eq: beta_j_i}
\beta_{j}^{(i)}=&\frac{1}{1+{\rm exp}\left(s_{j}^{(i)}\cosh^{-1}\left(\frac{1}{2\alpha_{j}^{(i)}}-1\right)\right)}\nonumber\\
&\quad\quad+s_{j}^{(i)}\alpha_{j}^{(i)}\cosh^{-1}\left(\frac{1}{2\alpha_{j}^{(i)}}-1\right), s_{j}^{(i)}\in \pm1.
\end{align}}With this parameterization embedded inside of \eqref{eq: dual_update}, we backpropagate through the objective function and take a gradient step with $\alpha$. Next, we clip all values of $\alpha$ to remain between $\underline{\alpha}$ and $\overline{\alpha}$, depending on whether the corresponding $\alpha$ is acting like an upper or lower affine bound. The full gradient-based verification routine for a sigmoid-based NN is given in Alg~\ref{alg:sigmoid_tightening}, which we refer to as $\alpha$-sig. The bounds $\overline x$, $\underline x$ for each activation function are treated as inputs.

\begin{algorithm}[tb]
\caption{Verifying sigmoid-based NNs with ``$\alpha$-sig"}
\label{alg:sigmoid_tightening}
\textbf{Input}: $\overline x$, $\underline x$ for each activation function\\
\textbf{Output}: $\tilde \gamma$: solution to \eqref{eq: dual_update}
\begin{algorithmic}[1] 
\STATE For all sigmoids, parallel compute $\underline{\alpha}$ and $\overline{\alpha}$ 
\STATE Initialize: $t=\infty$
\WHILE{$|{\tilde f}_r(\alpha,h(\alpha)-t|\ge \epsilon$}
\STATE Compute $t={\tilde f}_r(\alpha,h(\alpha))$
\FOR{NN Layers $i=L,L-1,...1,$}
\STATE Compute layer sign vector $s^{(i)}$ in \eqref{eq: sign_funcs}
\STATE Embed $s^{(i)}$ inside of $\beta$ via \eqref{eq: beta_j_i}
\ENDFOR

\STATE Back-propagate objective ${\tilde f}_r$ of \eqref{eq: dual_update} w.r.t. $\alpha$
\STATE Take a gradient ascent step with, e.g., Adam
\STATE Clip all $\alpha$ values between $\underline{\alpha}$ and $\overline{\alpha}$
\ENDWHILE
\STATE \textbf{return} tight lower bound $\tilde \gamma={\tilde f}_r(\alpha,\beta)$
\end{algorithmic}
\end{algorithm}

\section{Test Results}
In order to test the effectiveness of $\alpha$-sig and Alg.~\ref{alg:sigmoid_tightening}, we optimized over a range of randomly generated sigmoid-based NNs in two separate experiments. All NNs consisted of four dense layers with sigmoid activation functions, followed by a dense linear layer. We considered NNs containing 5, 10, 50, 100, 500, and 1000 neurons per layer. For each NN size, we generated and verified over 5 independent NN instantiations. The posed verification problem sought to minimize the sum of all outputs (i.e., the $c$ vector from \eqref{eq: c_eq} was set to the vector of all ones), and we assumed an allowable infinity norm perturbation of $\left\Vert x\right\Vert _{\infty}\le1$ (i.e., $-1\le x\le1$). To generate initial loose activation function bounds for all NN layers, we applied vanilla interval bound propagation via~\eqref{eq: ibp}, as reviewed in the appendix. We note that tighter bounds could potentially be achieved via IBP + backward mode LiRPA~\cite{xu2020autolirpa}, but we chose to utilize weaker IBP-based bounds to initialize $\alpha$-sig in order to highlight its effectiveness without strong initial activation function bounds.

In each test, we took 300 projected gradient steps in pursuit of solving \eqref{eq: dual_update}. We then benchmarked our results against $\alpha$-CROWN + auto-LiRPA~\cite{xu2021fast,xu2020autolirpa}. In order to fairly compare, we increased the default $\alpha$-CROWN iteration count by 3-fold, to 300 iterations. To avoid an early exit from CROWN (i.e., due to a successfully proved bound), we set the VNN-LIB~\cite{ferrari2022complete} verification metric to an arbitrarily high value $\Gamma$ (i.e., prove $f(x)\ge \Gamma,\;\forall x$). In order to compare the bounds proved via our $\alpha$-sig in Alg.~\ref{alg:sigmoid_tightening} vs $\alpha$-CROWN, we defined $\tau$:
\begin{align}\label{eq: tau}
\tau & =100\times\frac{\tilde{\gamma}-{\rm CROWN}_{{\rm bound}}}{|{\rm CROWN}_{{\rm bound}}|},
\end{align}
where $\tau$ represents the percent improvement or decline of our bound relative to $\alpha$-CROWN (positive $\tau$ means $\alpha$-sig provides a \textit{tighter} bound than $\alpha$-CROWN, while negative $\tau$ means the bound in looser). $\alpha$-sig was built in Julia, and all code is provided as supplementary material.

\begin{figure}[t]
\centering
\includegraphics[width=1\columnwidth]{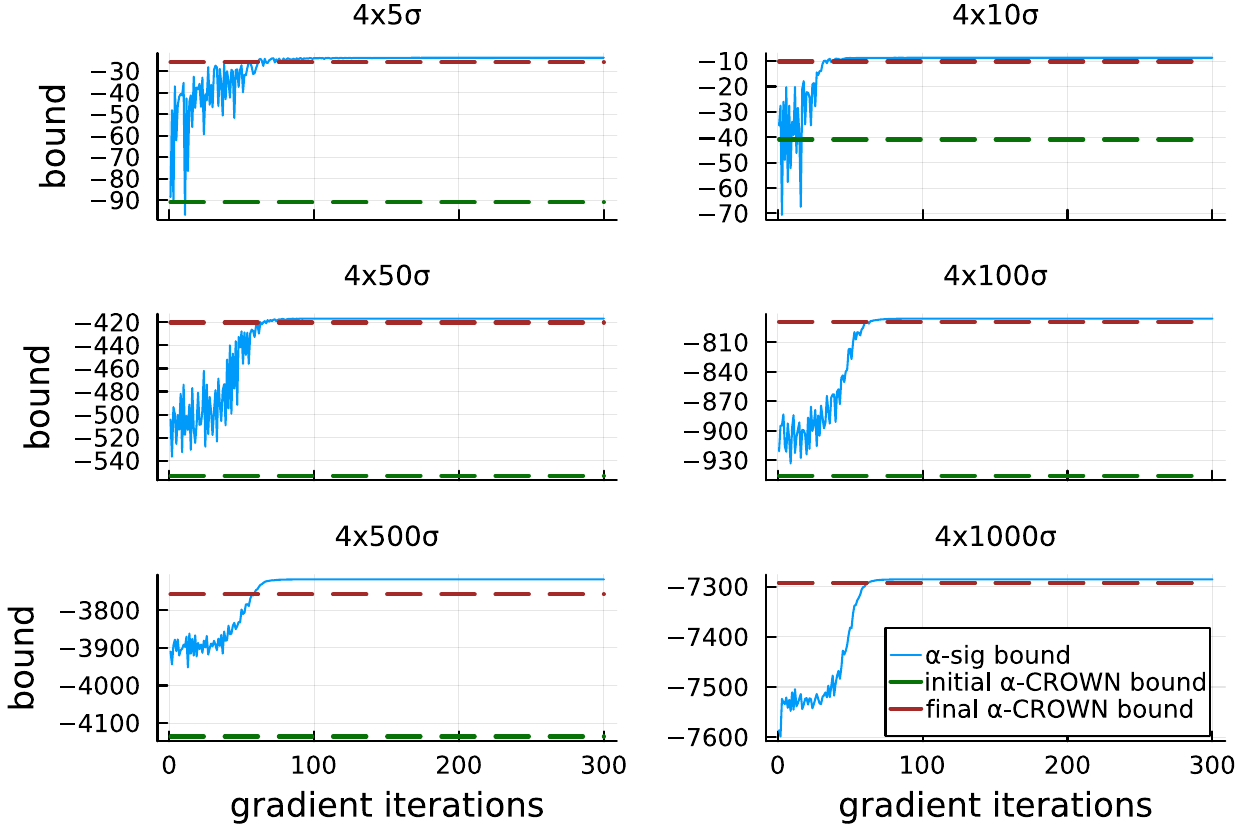}
\caption{Gradient ascent iterations of $\alpha$-sig across 6 NN model sizes. Test are associated with $\tau_1$ from Table \ref{tab:compareSolvers}. Initial and final (i.e., optimized) CROWN bounds are also shown.}
\label{ref:bound_race}
\end{figure}

\textbf{Experiment 1: Varying weight distributions.} Normally distributed weight and bias parameters tend to yield NNs whose sigmoid activation functions are always stuck on or off. In order to avoid this, we
we initialized all NN weights and biases via $W_i\!\sim\!\mathcal{N}(0,(2.5/j)^2)$ and $b_i\!\sim\!\mathcal{N}(0,0.25^2)$, where $j\in\{1,2,3,4,5\}$ represents the model index; thus, in this experiment, the weight parameter variances for each NN model progressively shrank (for a given model size). Results associated with this test are given in Table \ref{tab:compareSolvers}: in this table, the values $\tau_1$ through $\tau_5$ represent bound comparisons, à la \eqref{eq: tau}, across five independently generated models (five for each NN size). Clearly, $\alpha$-sig tended to provide marginally tighter bounds than $\alpha$-CROWN. Across all six NN sizes, the bound progressions for $\alpha$-sig are illustrated in Fig.~\ref{ref:bound_race}. In this figure, initial CROWN and fully optimized $\alpha$-CROWN bounds are superimposed for reference. 

In the upper-right portion of Table \ref{tab:compareSolvers}, $\alpha$-CROWN outperformed $\alpha$-sig. The reason why is apparent and interesting: in our experiments, $\alpha$-sig was initialized with fairly loose IBP-based primal bounds $\overline x$, $\underline x$ (see input to Alg.~\ref{alg:sigmoid_tightening}). As NN models shrink in size and weight parameter variances drop, auto-LiRPA's proclivity for primal bound solving seems to overtake $\alpha$-sig's optimal tightening of the sigmoid activation function. The benefit of $\alpha$-sig, however, is also in its speed. As demonstrated in Table \ref{tab:compareSolvers_time}, $\alpha$-sig can be up to two orders of magnitude faster\footnote{One of the trial times was removed due to CROWN error: ``Pre-activation bounds are too loose for BoundSigmoid".} than $\alpha$-CROWN, while still yielding better bounds in many cases.

\begin{table}
   \caption{Comparison of $\alpha$-sig vs $\alpha$-CROWN bounds (NN weight distribution variance shrink from $\tau_1\rightarrow \tau_5$).} 
   \label{tab:compareSolvers}
   \small
   \centering
   \begin{tabular}{c|cccccc}
   \toprule\toprule
   {$\;\,$\textbf{NN Size}} & $\tau_1$ & $\tau_2$ & $\tau_3$ & $\tau_4$ & $\tau_5$ \vspace{-0.35mm}\\  

   \midrule 
   {$4\times 5 \sigma$} & \textcolor{nicegreen}{+8.02}   & \textcolor{nicered}{-18.9}  & \textcolor{nicered}{-6.1}   & \textcolor{nicered}{-21.2}  & \textcolor{nicered}{-31.8}  \\
   \midrule 
   {$4\times 10 \sigma$} & \textcolor{nicegreen}{+14.6} & \textcolor{nicered}{-7.23}   & \textcolor{nicered}{-28.1}    & \textcolor{nicered}{-26.4} & \textcolor{nicered}{-24.0}    \\
   \midrule 
   {$4\times 50 \sigma$} & \textcolor{nicegreen}{+0.83} & \textcolor{nicegreen}{+0.51}  & \textcolor{nicegreen}{+0.23} & \textcolor{nicered}{-4.25} & \textcolor{nicered}{-9.63} \\
   \midrule 
   {$4\times 100 \sigma$} & \textcolor{nicegreen}{+0.40} & \textcolor{nicegreen}{+0.34}  & \textcolor{nicegreen}{+0.04} & \textcolor{nicered}{-0.52} & \textcolor{nicered}{-1.60} \\
   \midrule 
   {$4\times 500 \sigma$} & \textcolor{nicegreen}{+1.06} & \textcolor{nicegreen}{+0.13} & \textcolor{nicegreen}{+0.14} & \textcolor{nicegreen}{+0.10} & \textcolor{nicegreen}{+0.03}\\
      \midrule 
   {$4\times 1000 \sigma$} & \textcolor{nicegreen}{+0.09} & \textcolor{nicegreen}{+0.81} & \textcolor{nicegreen}{+0.11} & \textcolor{nicegreen}{+0.11} & \textcolor{nicegreen}{+0.06}\\
\bottomrule
   \end{tabular}
\end{table}

\begin{table}
   \caption{Mean $\alpha$-CROWN and $\alpha$-sig solve times.} 
   \label{tab:compareSolvers_time}
   \small
   \centering
   \begin{tabular}{c|ccc}
   \toprule\toprule
   {$\;\,$\textbf{NN Size}} & $\alpha$-CROWN & $\alpha$-sig & $\alpha$-sig speedup \vspace{-0.35mm}\\  

   \midrule 
   {$4\times 5 \sigma$} & 37.41 sec   & 0.10 sec & \textbf{360.4x} \\
   \midrule 
   {$4\times 10 \sigma$} & 37.77 sec & 0.105 sec & \textbf{358.4x}   \\
   \midrule 
   {$4\times 50 \sigma$} & 38.33 sec & 0.17 sec & \textbf{223.1x} \\
   \midrule 
   {$4\times 100 \sigma$} & 37.90 sec & 0.40 sec & \textbf{95.1x} \\
   \midrule 
   {$4\times 500 \sigma$} & 53.42 sec & 2.65 sec & \textbf{20.2x}\\
      \midrule 
   {$4\times 1000 \sigma$} & 82.90 sec & 9.98 sec & \textbf{8.3x}\\
\bottomrule
   \end{tabular}
\end{table}

\textbf{Experiment 2: Consistent weight distributions.} In this test, we initialized all NN weights and biases with consistent distributions: $W_i\!\sim\!\mathcal{N}(0,2.5^2)$ and $b_i\!\sim\!\mathcal{N}(0,0.25^2)$. Results from across five randomly initialized models are shown in Table~\ref{tab:compareSolvers_cd}. In contrast to the first experiment, $\alpha$-sig was able to reliably outperform $\alpha$-CROWN. In this case, the larger weight parameter variances seemed to cause inherently looser primal bounds, meaning $\alpha$-CROWN's advantage over $\alpha$-sig was lessened.

\begin{table}
   \caption{\centering Comparison of $\alpha$-sig vs $\alpha$-CROWN bounds (consistent distributions across models 1 through 5).} 
   \label{tab:compareSolvers_cd}
   \small
   \centering
   \begin{tabular}{c|cccccc}
   \toprule\toprule
   {$\;\,$\textbf{NN Size}} & $\tau_1$ & $\tau_2$ & $\tau_3$ & $\tau_4$ & $\tau_5$ \vspace{-0.35mm}\\  

   \midrule 
   {$4\times 5 \sigma$} & \textcolor{nicered}{-7.30}   & \textcolor{nicegreen}{+17.9}  & \textcolor{nicered}{-32.99}   & \textcolor{nicered}{-8.12}  & \textcolor{nicered}{-0.37}  \\
   \midrule 
   {$4\times 10 \sigma$} & \textcolor{nicegreen}{+4.6} & \textcolor{nicegreen}{+28.1}   & \textcolor{nicered}{-1.31}    & \textcolor{nicegreen}{+2.29} & \textcolor{nicered}{-2.66}    \\
   \midrule 
   {$4\times 50 \sigma$} & \textcolor{nicegreen}{+1.26} & \textcolor{nicegreen}{+0.49}  & \textcolor{nicegreen}{+0.64} & \textcolor{nicegreen}{+1.21} & \textcolor{nicegreen}{+0.95} \\
   \midrule 
   {$4\times 100 \sigma$} & \textcolor{nicegreen}{+0.47} & \textcolor{nicegreen}{+0.51}  & \textcolor{nicegreen}{+0.65} & \textcolor{nicegreen}{+0.70} & \textcolor{nicegreen}{+0.57} \\
   \midrule 
   {$4\times 500 \sigma$} & \textcolor{nicegreen}{+0.99} & \textcolor{nicegreen}{+0.98} & \textcolor{nicegreen}{+1.02} & \textcolor{nicegreen}{+0.97} & \textcolor{nicegreen}{+0.98}\\
      \midrule 
   {$4\times 1000 \sigma$} & \textcolor{nicegreen}{+0.13} & \textcolor{nicegreen}{+0.19} & \textcolor{nicegreen}{+0.09} & \textcolor{nicegreen}{+0.12} & \textcolor{nicegreen}{+0.12}\\
\bottomrule
   \end{tabular}
\end{table}

\section{Discussion and Conclusion}
Verifying over NNs containing S-shaped activation functions is inherently challenging. In this paper, we presented an explicit, differentiable mapping between the slope and y-intercept of an affine expression which tangentially bounds a sigmoid function. By optimizing over this bound's parameters in the dual space, our proposed convex relaxation of the sigmoid is maximally tight (i.e., a tighter element-wise relaxation of the sigmoid activation function does not exist). As explored in the test results section, however, our ability to fully exploit this tightness hinges on having good primal bounds $\overline x$, $\underline x$ for all activation functions. For example, given the bounds $\overline x=\infty$, $\underline x=-\infty$, our approach collapses to a useless box constraint. 

Even with relatively loose IBP-based activation function bounds, however, our proposed verification routine ``$\alpha$-sig" is able to ($i$) marginally outperform $\alpha$-CROWN in terms of bound tightness, and ($ii$) substantially outperform it in terms of computational efficiency. Future work will attempt to marry LiRPA/CROWN's excellent proclivity for activation function bounding with $\alpha$-sig's element-wise activation function tightening prowess. Future work will also extend the approach presented in this paper to other S-shaped activation functions, along with the multi-variate softmax function.

\appendix
\section{APPENDICES}
\section{Neural Network Mapping and Relaxation}
\label{sec:nn_mappng}
The NN mapping, whose output is transformed by a verification metric $c^T$, is stated below. At each sigmoid activation function, we also state the corresponding affine relaxation, where ${\rm d}(\alpha)$ diagonalizes an $\alpha$ vector into a diagonal matrix:
{\small\begin{subequations}\label{eq: nn_mapping}
\begin{align}
c^T{\rm NN}(x) & =c^T(W^{(L)}x^{(L)}+b^{(L)})\label{eq: c_eq}\\
x^{(L)} & =\sigma(\hat{x}^{(L-1)})\;\;:\textcolor{niceblue}{\;\; {\rm d}(\alpha^{(L-1)})\hat{x}^{(L-1)} + \beta^{(L-1)}}\\
\hat{x}^{(L-1)} & =W^{(L-1)}x^{(L-1)}+b^{(L-1)}\\
 & \vdots\nonumber\\
x^{(3)} & =\sigma(\hat{x}^{(2)})\;\;:\textcolor{niceblue}{\;\; {\rm d}(\alpha^{(2)})\hat{x}^{(2)} + \beta^{(2)}}\\
\hat{x}^{(2)} & =W^{(2)}x^{(2)}+b^{(2)}\\
x^{(2)} & =\sigma(\hat{x}^{(1)})\;\;:\textcolor{niceblue}{\;\; {\rm d}(\alpha^{(1)})\hat{x}^{(1)} + \beta^{(1)}}\\
\hat{x}^{(1)} & =W^{(1)}x^{(1)}+b^{(1)}\\
x^{(1)} & =x.
\end{align}
\end{subequations}}We may now relax the sigmoid activation functions, replacing them with their affine bounds. To determine if each scalar $\alpha$, $\beta$ pair are chosen to be an upper bound or a lower bound, we need to determine the coefficient in front of the corresponding activation function. To do this, we move backward through the NN, defining vectors $s^{(i)}$ of sign values corresponding to the signs of the entries in the argument:
\begin{subequations}\label{eq: sign_funcs}
\begin{align}
s^{(L)} & ={\rm sign}(c^{T}W^{(L)})\\
s^{(L-1)} & ={\rm sign}(c^{T}W^{(L)}{\rm d}(\alpha^{(L-1)})W^{(L-1)})\\
 & \;\; \vdots\nonumber\\
s^{(2)} & ={\rm sign}(c^{T}W^{(L)}{\rm d}(\alpha^{(L-1)})W^{(L-1)}...\nonumber\\
 & \quad\quad\quad\quad\quad\quad{\rm d}(\alpha^{(2)})W^{(2)})\\
s^{(1)} & ={\rm sign}(c^{T}W^{(L)}{\rm d}(\alpha^{(L-1)})W^{(L-1)}...\nonumber\\
 & \quad\quad\quad\quad\quad\quad{\rm d}(\alpha^{(2)})W^{(2)}{\rm d}(\alpha^{(1)})W^{(1)}).
\end{align}
\end{subequations}

\section{Slope and Intercept Relation Derivation}
Starting from \eqref{eq: tangent}, the sigmoid derivative can be written via $\sigma'(x) = \sigma(x)(1-\sigma(x))$. Setting this equal to the upper affine bound slope, we have
\begin{subequations}
\begin{align}
\alpha & =\frac{1}{1+e^{-x}}\left(1-\frac{1}{1+e^{-x}}\right)\\
 & =\frac{e^{-x}}{\left(1+e^{-x}\right)^{2}}.
\end{align}
\end{subequations}
Multiplying through by the expanded right-side denominator, we may solve for the primal variable $x$ explicitly:
\begin{align}
\alpha\left(1+e^{-2x}+2e^{-x}\right) & =e^{-x}\\
\alpha\left(e^{x}+e^{-x}+2\right) & =1\\
2\alpha\left(\cosh(x)+1\right) & =1\\
x & =\pm\cosh^{-1}\left(\frac{1}{2\alpha}-1\right).
\end{align}
Since $\cosh(x)=\cosh(-x)$, the solution for $x$ is non-unique, which is an inherent consequence of the symmetry of the sigmoid function, where exactly two points in the sigmoid manifold will map to the same slope. Notably, however, the affine upper bound will map to positive solutions of $x$, i.e., in the region where the sigmoid function is concave, and the affine lower bound will map to negative values of $x$, i.e., in the region where the sigmoid function is convex:
\begin{align}
x & =\cosh^{-1}\left(\frac{1}{2\alpha_{u}}-1\right)\label{eq: primal1}\\
x & =-\cosh^{-1}\left(\frac{1}{2\alpha_{l}}-1\right).\label{eq: primal2}
\end{align}
Reorganizing the intersection equation \eqref{eq: intersect}, such that $\beta =1/(1+e^{-x})-\alpha x$, we may plug the primal solutions \eqref{eq: primal1}-\eqref{eq: primal2} in for $x$. This yields explicit expressions for the intercept points ($\beta_{u}$, $\beta_{l}$) as functions of the slopes ($\alpha_{u}$, $\alpha_{l}$):
\begin{align}
\beta_{u} & =\frac{1}{1+e^{-\cosh^{-1}\left(\frac{1}{2\alpha_{u}}-1\right)}}\!-\!\alpha\cosh^{-1}\!\left(\frac{1}{2\alpha_{u}}-1\right)\label{eq: beta_u}\\
\beta_{l} & =\frac{1}{1+e^{+\cosh^{-1}\left(\!\frac{1}{2\alpha_{l}}-1\!\right)}}\!+\!\alpha\cosh^{-1}\left(\!\frac{1}{2\alpha_{l}}-1\!\right).\label{eq: beta_l}
\end{align}

\section{Computing Maximum Slope Limits}
\begin{figure}[t]
\centering
\includegraphics[width=0.99\columnwidth]{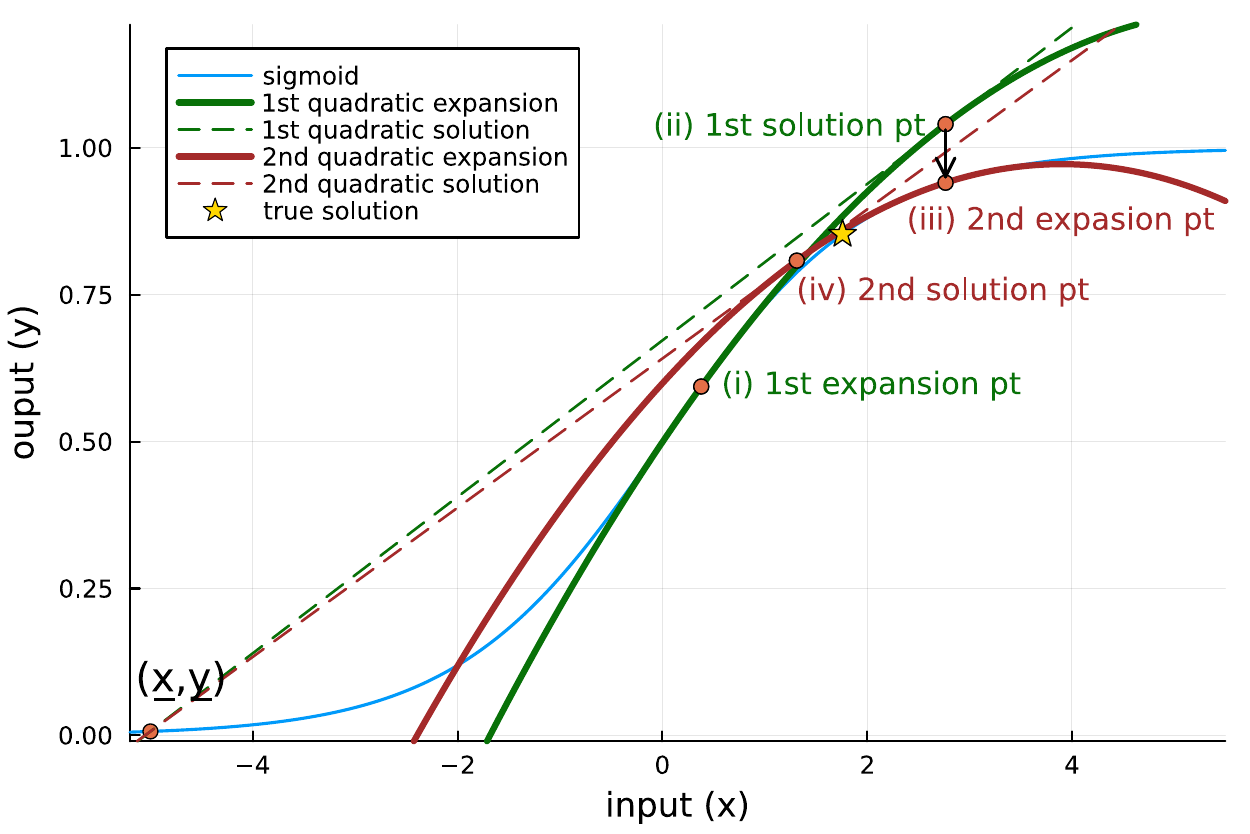}
\caption{Sequential quadratic expansions and solutions.}
\label{ref:quadratic_expansions}
\end{figure}

In order to compute the maximum slope values $\overline{\alpha}_{l}$ and $\overline{\alpha}_{u}$, as depicted in Fig.~\ref{sig_4_bounds}, we use a sequential numerical routine which iteratively solves a quadratic expansion of the associated problem. Consider the following system of equations, with unknown variables $\alpha$, $\beta$, and $\hat x$ (intercept point):
\begin{align}
\underline{y} & =\alpha\underline{x}+\beta && \text{anchor point}\label{eq: yaxb}\\
\sigma(\hat{x}) & =\alpha\hat{x}+\beta && \text{sigmoid intersection at } {\hat x}\\
\sigma(\hat{x})' & =\alpha && \text{match sigmoid slope at } {\hat x},\label{eq: alpha_soln}
\end{align}
where the known ``anchor point" $(\underline{x}, \underline{y})$ is depicted in Fig.~\ref{ref:quadratic_expansions}. Despite its similarity to \eqref{eq: affine_sigmoid_solution}, this system does not have a closed-form solution (i.e., it will result in a single nonlinear equation, similar to \eqref{eq: beta_f_alpha}, but with $\beta$ replaced by an expression for $\alpha$). In order to efficiently solve this system of equations, we perform a single quadratic expansion of the sigmoid activation function~\cite{agarwal2018}:
\begin{align}
\sigma(\hat{x})\approx \tilde{\sigma}(\hat{x}) & \triangleq\sigma_{0}+\sigma'_{0}(\hat{x}-\hat{x}_{0})+\frac{1}{2}\sigma''_{0}(\hat{x}-\hat{x}_{0})^{2},
\end{align}
where $\sigma_{0}\triangleq\sigma(x)|_{{\hat x}_{0}}$, etc., is used for notational simplicity. Collecting like powers of $\hat{x}$, the quadratic expansion yields
{\small\begin{align*}
\tilde{\sigma}(\hat{x}) & =\sigma_{0}+\sigma'_{0}\hat{x}-\sigma'_{0}\hat{x}_{0}+\frac{1}{2}\sigma''_{0}(\hat{x}^{2}+\hat{x}_{0}^{2}-2\hat{x}\hat{x}_{0})\\
 & =\sigma_{0}+\sigma'_{0}\hat{x}-\sigma'_{0}\hat{x}_{0}+\frac{1}{2}\sigma''_{0}\hat{x}^{2}+\frac{1}{2}\sigma''_{0}\hat{x}_{0}^{2}-\sigma''_{0}\hat{x}\hat{x}_{0}\\
 & =\underbrace{\left(\sigma_{0}-\sigma'_{0}\hat{x}_{0}+\frac{1}{2}\sigma''_{0}\hat{x}_{0}^{2}\right)}_{c_{0}}+\underbrace{\left(\sigma'_{0}-\sigma''_{0}\hat{x}_{0}\right)}_{c_{1}}\hat{x}+\underbrace{\frac{1}{2}\sigma''_{0}}_{c_{2}}\hat{x}^{2}.
\end{align*}}
The updated system of equations is now quadratic:
\begin{align}\underline{y} & =\alpha\underline{x}+\beta &  & \text{anchor point}\\
c_{2}\hat{x}^{2}+c_{1}\hat{x}+c_{0} & =\alpha\hat{x}+\beta &  & \text{approx intersection}\label{eq: approx_int}\\
2c_{2}\hat{x}+c_{1} & =\alpha &  & \text{approx slope match}.
\end{align}
In this system, we may eliminate the $\alpha$ and $\beta$ terms by setting $\alpha = 2c_{2}\hat{x}+c_{1}$ and $\beta=\underline{y}-\alpha\underline{x}$, which yields $\beta=\underline{y}-\left(2c_{2}\hat{x}+c_{1}\right)\underline{x}$. Plugging these into \eqref{eq: approx_int}, a single quadratic equation emerges:
\begin{align*}
c_{2}\hat{x}^{2}+c_{1}\hat{x}+c_{0} & =\left(2c_{2}\hat{x}+c_{1}\right)\hat{x}+\underline{y}-\left(2c_{2}\hat{x}+c_{1}\right)\underline{x}.
\end{align*}
We reorganize these terms into a standard quadratic form:
\begin{align}\label{eq: quad_d}
0=\underbrace{\left(c_{2}\right)}_{d_{2}({\hat x}_0)}\hat{x}^{2}+\underbrace{\left(-2c_{2}\underline{x}\right)}_{d_{1}({\hat x}_0)}\hat{x}+\underbrace{\left(\underline{y}-c_{1}\underline{x}-c_{0}\right)}_{d_{0}({\hat x}_0)},
\end{align}
where coefficients $d_{2}$, $d_{1}$, $d_{0}$ are written as functions of the expansion point  ${\hat x}_0$. We use the quadratic formula to analytically solve \eqref{eq: quad_d}. Multiple iterations of ($i$) updating the $d$ coefficients and then ($ii$) re-solving \eqref{eq: quad_d} via quadratic formula yields rapidly converging solutions to the original system~\eqref{eq: yaxb}-\eqref{eq: alpha_soln}. Due to the analytical exactness of the quadratic formula, this routine converges faster than Newton iterations at a similar computational expense. Two iterations of this routine are depicted in Fig.~\ref{ref:quadratic_expansions}, where the second solution falls very close to the true solution (which future iterations converge to). Once the solution ${\hat x}^*$ is found, we recover $\alpha$ via $\alpha =\sigma'(\hat{x}^{*})$. The procedure is applied to find the maximum upper affine bound slope in Fig.~\ref{ref:quadratic_expansions}; an identical procedure is applied to find the maximum \textit{lower} affine bound slope.

\section{Interval Bound Propagation}
In order to propagate input bounds ${\underline x}\le x\le {\overline x}$ through a NN layer $y=\sigma (W_i x + b_i)$, we employ interval bound propagation~\cite{gowal2019effectivenessintervalboundpropagation}. Defining mean $x_{\mu}=\tfrac{1}{2}(\overline{x}+\underline{x})$ and deviation $x_{\sigma} =\tfrac{1}{2}(\overline{x}-\underline{x})$ vectors, the output bound vectors are given as
\begin{subequations}\label{eq: ibp}
\begin{align}
\overline{y} & =\sigma\left(\left|W_{i}\right|x_{\sigma}+W_{i}x_{\mu}+b_{i}\right)\\
\underline{y} & =\sigma\left(-|W_{i}|x_{\sigma}+W_{i}x_{\mu}+b_{i}\right),
\end{align}
\end{subequations}
where $\sigma(\cdot)$ can be any element-wise monotonic activation function~\cite{gowal2019effectivenessintervalboundpropagation}.

\section{Acknowledgments}
The authors thank Dr.~Amrit Pandey, who suggested the idea of using sequential quadratic expansions of the sigmoid function in order to solve \eqref{eq: yaxb}-\eqref{eq: alpha_soln}.

\bibliography{aaai25}
\end{document}